\documentclass{article}

    \PassOptionsToPackage{numbers, compress}{natbib}

    \usepackage[preprint]{neurips_2024}

\usepackage[utf8]{inputenc} 
\usepackage[T1]{fontenc}    
\usepackage{hyperref}       
\usepackage{url}            
\usepackage{booktabs}       
\usepackage{amsfonts}       
\usepackage{nicefrac}       
\usepackage{microtype}      
\usepackage{xcolor}         

\usepackage{xspace}
\usepackage{amssymb}
\usepackage{mathtools}
\usepackage{adjustbox}
\usepackage{multirow}
\usepackage{todonotes}
\usepackage{geometry}
\usepackage{array}
\usepackage{capt-of}
\usepackage{graphicx}

\newcommand{\method}{CROME\xspace}

\title{\method: Cross-Modal Adapters \\ for Efficient Multimodal LLM}

\author{%
  Sayna Ebrahimi, Sercan \"{O}. Ar{\i}k, Tejas Nama, Tomas Pfister\\
Google Cloud AI Research \\
\texttt{\{saynae, soarik, tejasnama, tpfister\}@google.com}
}

\begin{document}

\maketitle

\begin{abstract}
Multimodal Large Language Models (MLLMs) demonstrate remarkable image-language capabilities, but their widespread use faces challenges in cost-effective training and adaptation. 
Existing approaches often necessitate expensive language model retraining and limited adaptability. 
Additionally, the current focus on zero-shot performance improvements offers insufficient guidance for task-specific tuning. 
We propose CROME, an efficient vision-language instruction tuning framework. It features a novel gated cross-modal adapter that effectively combines visual and textual representations prior to input into a frozen LLM. This lightweight adapter, trained with minimal parameters, enables efficient cross-modal understanding. Notably, CROME demonstrates superior zero-shot performance on standard visual question answering and instruction-following benchmarks. Moreover, it yields fine-tuning with exceptional parameter efficiency, competing with task-specific specialist state-of-the-art methods. CROME demonstrates the potential of pre-LM alignment for building scalable, adaptable, and parameter-efficient multimodal models. 
\end{abstract}

\begin{figure}[t]
\begin{center}
  \includegraphics[width=\textwidth]{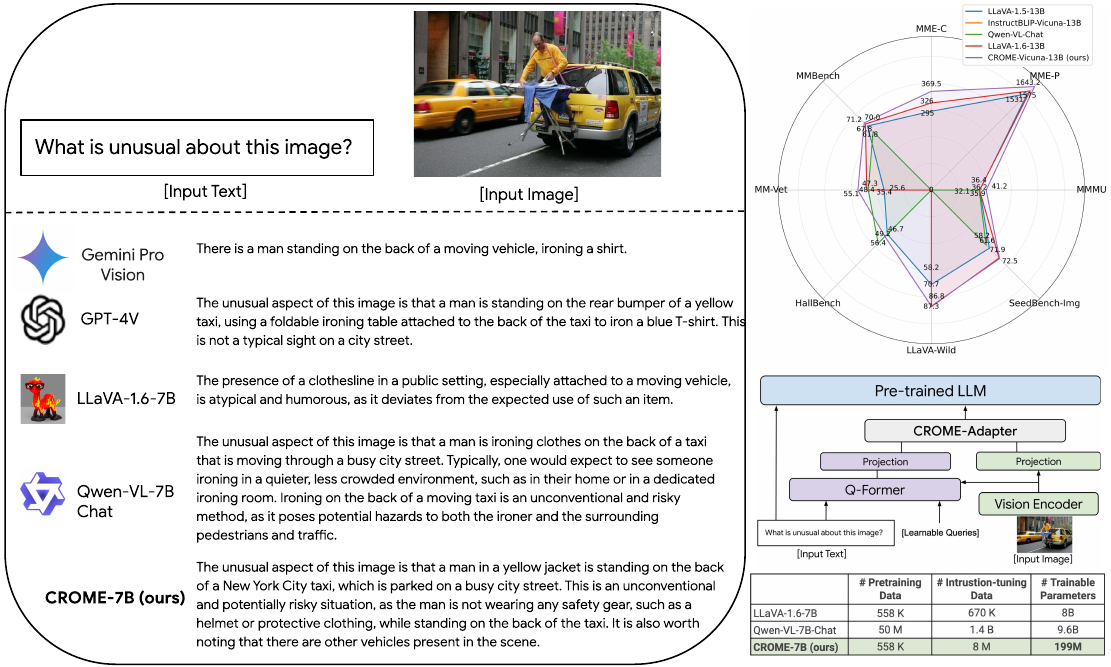}
\end{center}
  \caption{\method achieves state-of-the-art results on 6 MLLM benchmarks (top right) with its unique pre-LM cross-modal adapter (middle right). Bottom Right: Training data and parameter comparisons.
  Left: A qualitative example on the unusual "ironing man" image.} 
\label{fig:teaser}
\end{figure}

\begin{figure}[t]
\begin{center}
  \includegraphics[width=\textwidth]{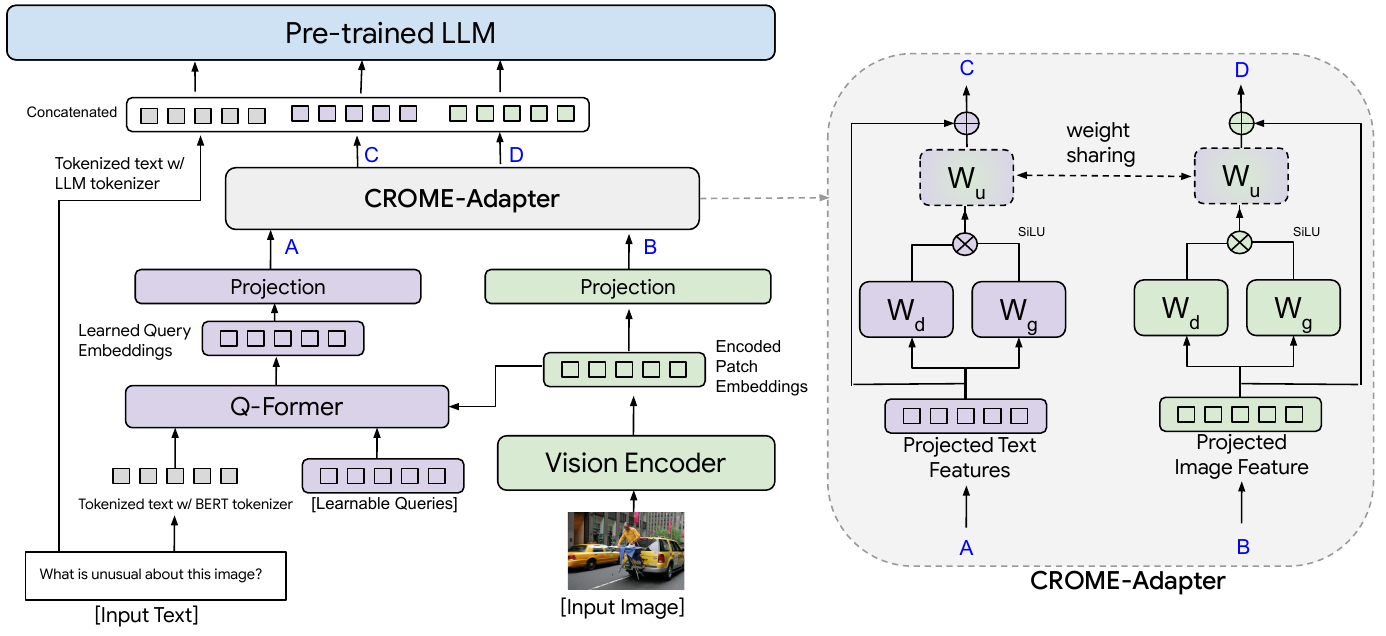}
\end{center}
  \caption{\small{\textbf{Overview of our \method model architecture with \method-Adapter.} \method takes both image and text as input to generate output text autoregressively. The text input query is encoded by the Q-Former, which utilizes learnable queries to effectively represent instruction-aware visual features. These are then processed by a projection layer. The image input is encoded by a vision encoder and its patch embeddings are used in both the Q-Former's cross-attention layers and a projection layer. Within the cross-modal adapter, the projected image and text features undergo individual down-projection using a gated linear unit ([see~\ref{sec:crossadapter}]). They are then up-projected through a weight sharing linear layer. Finally, the cross-modal adapter outputs for text and image are concatenated with the tokenized question and fed into the LLM to obtain the text output.}}
\label{fig:crome}
\end{figure}

\section{Introduction}
\label{submission}

Recent advancements in Multimodal Large Language Models (MLLMs) have yielded impressive breakthroughs across many scenarios, particularly in  vision-language learning. Notably, OpenAI's GPT-4v \cite{gpt4v} and Google's Gemini Pro Vision \cite{gemini} demonstrate exceptional performance for tasks like image captioning and visual question answering. Such commercial models are typically available through prediction-only APIs that limit their wider adaptation and customization.
On the other hand, there are notable open-sourced vision-language models, including LLaVA~\cite{llava}, InstructBLIP~\cite{instructblip}, Qwen-VL~\cite{qwenvl}, and BLIVA~\cite{bliva}. These are often built using instruction tuning to improve multimodal capabilities of LLMs. Their success has been shown to depend on large-scale training data (can be hundreds of millions~\cite{blip2,instructblip,idefics,bliva} to over a billion~\cite{qwenvl}), often yielding high training costs\cite{llava,mplugowl,kosmos,qwenvl}. 
Efficient methods for building large vision-language models from available vision-only or text-only pretrained models, as well as tuning them for target multimodal use cases, remain everlasting challenges. 

The extensive parameter counts involved in training image encoders and language models is one key challenge behind high computational costs. 
While retraining the LLMs with multimodal data helps align visual and textual tokens~\cite{llava,mplugowl,kosmos,flamingo}, it can come with the risk of undermining the pretrained LLMs' reasoning capabilities~\cite{lavin,qwenvl}. 
Furthermore, as the variety of LLMs continues to grow, a retraining approach hinders their potential for `plug-and-play integration' within multimodal frameworks. 
We propose that pre-aligning visual and textual tokens before LLM ingestion offers a more flexible, efficient, and scalable strategy that remains under-explored. 
Our empirical findings demonstrate that the way image and text representations are adapted for LLM compatibility can significantly affect multimodal understanding and reasoning. 
Most models still rely on simplistic linear projections before token concatenation~\cite{llava, shikra}. 
While BLIP-2~\cite{blip2} and InstructBLIP~\cite{instructblip} partially address the need for efficient cross-modal learning by introducing a Query Transformer (Q-Former)~\cite{instructblip,bliva}, they still face challenges. 
Their pretraining remains computationally expensive (approximately 100 GPU hours on >100M image-text pairs), and fine-tuning for specific domains can be parameter-inefficient.

For real-world applications, improving MLLMs to excel in specific tasks is essential, extending their value beyond zero-shot scenarios. 
While zero-shot performance results showcase their potential to handle diverse tasks without training data, there is also a significant opportunity to maximize their effectiveness for cases where data for specific downstream tasks are available. 
In such cases, the ability to implement efficient and adaptable tuning strategies becomes crucial. 
While some MLLMs, such as LLaVA, suggest parameter-efficient tuning techniques like LoRA~\cite{lora} or selectively training the image-language projector layer, the optimal implementation and generalizability of these across diverse tasks and datasets warrant further exploration. 
Developing effective, cost-efficient and flexible tuning strategies would not only unlock the full potential of MLLMs for targeted applications but also ensure improved and robust performance beyond zero-shot benchmarks.

In this paper, we propose \method, a vision-language training framework featuring a vision encoder, query Transformer, and a novel gated cross-modal adapter depicted in~\ref{fig:teaser}. The proposed adapter unifies vision and language representations prior to LLM input, promoting superior cross-modal understanding while maintaining parameter efficiency by keeping both the LLM and vision encoder frozen. Our lightweight cross-modal fusion unit effectively learns cross-modal relations, making fine-tuning \method remarkably straightforward. Our contributions can be summarized as:

\begin{itemize} 
\item We present \method, a novel vision-language learning framework featuring a lightweight gated cross-modal adapter (\method-Adapter) which is used for aligning visual and textual tokens before LLM input for multimodal learning. This avoids costly LLM training and maintains generalization on text understanding and reasoning tasks. 

\item \method introduces an effective, cost-efficient and flexible fine-tuning strategy to maximize MLLM effectiveness with availability of data from specific downstream tasks. The \method-Adapter's design enables both cross-modal understanding and parameter-efficient fine-tuning, as only the adapter is trained during adaptation.

\item We evaluate \method's performance on a diverse set of MLLM benchmarks for zero-shot and supervised fine-tuning scenarios and show that \method outperforms the state-of-the-art open-source baselines on 6/8 benchmarks. Training only the cross-modal adapter (nearly 5M parameters), we demonstrate that \method outperforms state-of-the-art methods specifically tailored for individual tasks.

\end{itemize}

\section{Related Work}

\subsection{Multimodal large language models (MLLMs)}
Vision-language models (VLMs) are often based on aligning image and text features in a unified embedding space, as proposed by the notable works CLIP~\cite{clip} and ALIGN~\cite{align}, followed by the subsequent ones \cite{coca,pali,blip,ofa}. This alignment is achieved through contrastive learning objectives applied to extensive image-text pair datasets. VLMs achieve strong zero-shot and few-shot performance, showcasing significant generalization abilities across a range of downstream tasks. Benefiting from existing LLMs and vision encoders of VLMs as the visual backbone (\textit{e.g.} CLIP's ViT encoder), recent MLLMs \cite{llava,minigpt,gemini,gpt4,instructblip,blip2,kosmos,bliva,instructblip,qwenvl,shikra} achieve even greater visual perception, understanding, and reasoning abilities. Flamingo~\cite{flamingo} establishes a connection between the vision encoder and LLMs using a Perceiver Resampler, showcasing remarkable few-shot performance.  BLIP-2~\cite{blip2} introduces a Q-former to align visual features with OPT~\cite{opt} and FLAN-T5~\cite{flant5}. MiniGPT-4~\cite{minigpt4} connects a ViT and Q-former with Vicuna~\cite{vicuna} as the LLM using a linear projector. Recently using large-scale paired instruction-image data has become a popular technique to adapt LLMs to answer questions regarding a given image~\cite{llava,minigpt,gemini,gpt4,instructblip,blip2,kosmos}. These approaches often involve retraining the LLM and/or the vision encoder, which can be computationally demanding. As LLMs continue to evolve and gain capabilities, it becomes increasingly important for MLLMs to leverage the existing strengths of LLMs without requiring retraining. This can avoid the risk of catastrophic forgetting, where retraining could potentially degrade the LLM's core natural language processing (NLP) abilities~\cite{lavin, qwenvl}.  Qwen-VL\cite{qwenvl} avoids this by adding text-only data to their pretraining and instruction-tuning. Existing methods that avoid LLM or vision encoder retraining, such as InstructBLIP and BLIVA \cite{instructblip, bliva}, primarily focus on learning cross-modal interactions within a Q-Former module, which is then retrained during supervised fine-tuning. While \method also utilizes a Q-Former, it introduces a novel, lightweight gated cross-modal adapter. 
This adapter differentiates our approach from BLIVA and InstructBLIP in two major ways:
(i) it enhances cross-modal understanding, serving as an additional unit to further refine cross-modal interactions; and 
(ii) it acts as the sole trainable component during task-specific fine-tuning, maximizing tuning efficiency while preserving the existing instruction-aware capabilities in the Q-Former and the LLM.

\subsection{Parameter Efficient Tuning}
As LLMs grow in size, parameter-efficient tuning (PET) becomes essential for reducing memory and for cost-efficient training. PET involves selectively adding or adjusting a small number of parameters within a pretrained model for task-specific adaptation. Early PET methods focused on either language~\cite{petl,unifiedpetl,adapterfusion,lora} or vision data~\cite{residualadapters,Rebuffi2018EfficientPO,Chen2022AdaptFormerAV,repadapter}. Extending beyond unimodal learning, vision and language adapters were introduced for smaller models~\cite{vlpet,coda,vladapter}.  Approaches like MultiModal-GPT~\cite{multimodalgpt} and others~\cite{du2022glm,mplugowl} utilized LoRA~\cite{lora} within architectures like Flamingo~\cite{flamingo}. Similarly, LLaMA-adapter employed prefix tuning~\cite{prefix} with image embeddings as prefix tokens. Recent methods like LaVIN~\cite{lavin} and PILL~\cite{pill} focus on adapting LLMs for multimodal instructions. LaVIN employs the AdaMix adapter~\cite{adamix} in LLaMA and RepAdapter~\cite{repadapter} in ViT in an end-to-end manner. However, these adapters are integrated within Transformer blocks, and their performance generalization to other language models remains unexplored. 
\method distinguishes itself with a modular cross-modal adapter that works with encoder-decoder and decoder-only LLMs. During large-scale instruction tuning, both the adapter and Q-Former are trained. Crucially, for supervised fine-tuning on smaller datasets, the adapter becomes the sole trainable component. This approach yields more flexibility and efficiency given the wide range of adaptation scenarios, while protecting the LLMs' existing capabilities.

\section{\method: a Cross-modal adapter based MLLM}\label{sec:crossadapter}

\method is a multimodal LLM framework which receives image and text as the multimodal input and generates text in an autoregressive manner. 
We first introduce \method's architecture, focusing on the proposed adapter modules for superior MLLM results. Then, we explain the applied multimodal training methods: pretraining, instruction tuning, and the optional task-specific tuning. 

\subsection{Model architecture}

\method is composed of a pretrained frozen LLM, frozen vision encoder, and a query Transformer components, as shown in ~\ref{fig:crome}.
The projected image patch and query embeddings are passed to a cross modal adapter before getting concatenated with text embeddings and fed into the LLM. 
Note that although the high level architecture resembles the InstructBLIP family of MLLMs, this important aspect of how visual-text are processed before inputting them into LLMs differentiates \method. 
Below, we go through the details of each component:\\

\noindent{\textbf{Vision encoder.}} We utilize a vision encoder to extract image features, which are then processed by a linear projection layer and the Q-Former. During pretraining and instruction tuning, we keep the vision encoder itself frozen and maintain its pretrained visual representations, in order to obtain low-cost and parameter-efficient training.  Only the associated projection layer is trained during these stages. For details on image preprocessing, please refer to the Appendix.\\

\noindent{\textbf{Large language model.}} To ensure low-cost and parameter-efficient training, as well as improved generalization with limited tuning data, we propose keeping the LLM entirely frozen during all stages of training of the \method. This enables utilizing both decoder-only and encoder-decoder type model architectures as the LLMs. \\

\noindent{\textbf{Query Transformer (Q-Former).}} We employ a Q-Former architecture, in which queries interact with each other through self-attention and with frozen image features through cross-attention, which is inserted every other Transformer block. We initialize it with InstructBLIP's model Q-Former and all its 188M parameters are trained during the instruction-tuning phase, which only has nearly 20\% and 2.6\% of the vision encoder and the LLM parameters, respectively, which ensures the parameter efficiency during large-scale instruction tuning.\\

\noindent{\textbf{\method adapter}}  Inspired by the concept of adapter networks \cite{adapters}, which introduce parameter efficiency, we propose a lightweight cross-modal module within \method. Crucially, unlike typical adapter placement after feed-forward and self-attention layers in Transformers, this module facilitates the fusion of textual and visual representations before they enter into the LLM. This pre-LLM fusion offers a potential advantage for aligning different modalities for optimal understanding within the LLM. During fine-tuning, the cross-modal adapters are the only trainable components, enabling remarkably-efficient adaptation of \method allowing it to adapt to new tasks without extensive retraining of the core LLM.

As shown in~\ref{fig:crome}, we use a conventional bottleneck structure~\cite{adapters} with down-projection and up-projection units and skip connections. This design allows for efficient processing of high-dimensional input features. We use a modality-specific down-sampling unit for vision and text branches, where in each of them an input $d$-dimensional feature vector is projected to a smaller dimension, $m$. Inspired by the success of gated linear units in feedforward layers from Transformers~\cite{shazeer2020glu,gatedconv}, in the down-projection unit we use the component-wise product of two linear transformations named as $\textbf{W}_d\in\mathbb{R}^{d\times m}$ and $\textbf{W}_g\in\mathbb{R}^{d\times m}$ where the input one of which is sigmoid-activated~\cite{gelu,elfwing2018sigmoid}. This gating mechanism helps the adapter control the flow of information, potentially emphasizing the most useful and relevant multimodal relationships.

For each down-projection unit, given an input text or image feature embedding of size $x\in\mathbb{R}^d$, the output is mapped as:
\begin{equation}\label{eq:gating}
\textbf{z}(x) = \text{SiLU}(x \textbf{W}_d) \otimes x \textbf{W}_g,
\end{equation}
where $\text{SiLU}$ is Sigmoid Linear Unit function~\cite{gelu,elfwing2018sigmoid}. On the other hand, the up-projection unit uses a weight-sharing mechanism between the two modalities where the $m$-dimensional vector $\textbf{z}\in\mathbb{R}^{m}$ is projected back to $d$ input dimensions via $\textbf{W}_u\in\mathbb{R}^{m\times d}$, in order to better encourage learning of cross-modal relations. Overall, the output of each branch of the cross-modal adapter can be formulated as:
\begin{equation}
\text{\method-Adapter}(x, \textbf{W}_{d},\textbf{W}_{g}, \textbf{W}_{u}) = x + \textbf{z} \textbf{W}_{u} 
\end{equation}
Finally, the inputs to the LLM are formed by concatenating the tokenized text, the output of the text branch within the \method-Adapter, and the output of the vision branch of the \method-Adapter (see \ref{fig:crome}).

\begin{figure*}[t]
\begin{center}
  \includegraphics[width=\textwidth]{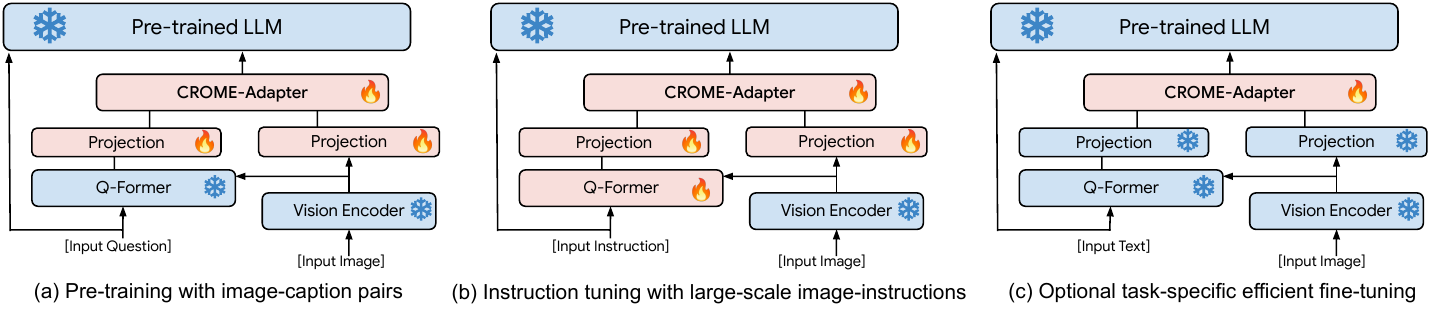}
\end{center}
  \caption{Overview of \method training stages. Blue indicates frozen components, and red indicates trainable components. (a) Pretraining: \method-Adapter and projection layers are trained on image-caption pairs. (b) Instruction-tuning: Q-Former, \method-Adapter, and projection layers are trained on diverse image-instruction datasets. (c) Task-specific fine-tuning: \method-Adapter facilitates efficient training on task-specific data.}
\label{fig:crome-training}
\end{figure*}

\subsection{Training \method}
In this section, we describe different training processes for \method: (a) pretraining with image-caption pairs followed by (b) instruction tuning with image-instructions on a variety of tasks, and (c) optional task-specific efficient fine-tuning which is used if data is available for a specific target task to optimize \method's task-specific performance. Throughout these stages, we use next token prediction as the training objective where the LLM predicts the next word conditioned on previous multimodal visual and text tokens. This encourages the model to accurately generate subsequent tokens based on the context of preceding tokens. Figure \ref{fig:crome-training} provides a visual representation of the training stages and trainable model components, described in detail below.
 \\

\noindent{\textbf{Pretraining.}} Our approach begins with a pretraining phase designed to align modalities within the projection layers. As shown in~\ref{fig:crome-training}(a), during this stage, we train the image and text projection layers alongside the cross-modal adapter. The remaining model layers are kept frozen. 
\\

\noindent{\textbf{Instruction tuning.}} At this stage, the model is refined to follow instructions accurately. We utilize a diverse set of image-instruction pairs to train the model to answer specific queries about images, extending its abilities beyond the image captioning learned during pretraining. During instruction tuning, we train the Q-Former, projection layers, and \method-adapter parameters. This enables the model to efficiently learn instruction-aware queries, facilitated by the cross-modal interaction between image embeddings and queries within the Q-Former (see ~\ref{fig:crome-training}(b)). The result of this instruction tuning is a model capable of strong zero-shot performance on visual question-answering benchmarks.
\\

\noindent{\textbf{Optional task-specific fine-tuning.}} When additional task-specific data (often smaller scale than the previous stage) is available, this step further optimizes \method's performance at the target task. The \method-Adapter allows for efficient fine-tuning by limiting the number of trainable parameters to approximately 5M (see \ref{fig:crome-training}
(c)). Besides low cost task-specific tuning, such parameter efficiency yields constitutes an effective mechanism to prevent overfitting, a commonly-observed challenge with small amount of task-specific data.

\section{Experiments}

In this section, we first discuss the datasets used to train \method, followed by implementation details including model architecture and training parameters. We note that our model and data will be publicly available.
Finally, we outline the benchmarks used to evaluate \method's performance.

\subsection{Datasets for Training \method}

As for the pretraining (PT) dataset, we use the LLaVA Pretrain LCS-558K~\cite{llava} which is a filtered subset of LAION/CC/SBU dataset as image-caption pairs consistently across all experiments. 
As for the instruction tuning (IT) dataset, we consider three different options for different experiments to highlight different outcomes:
\begin{itemize}
    
\item IT Dataset 1 (665K samples), used for evaluating pre-LM modality alignment: To demonstrate the effectiveness of \method's pre-LM input alignment units compared to LLM retraining used in models such as LLaVA, we use the same 665K instruction-image pairs as LLaVA1.5~\cite{llava}~\footnote{LLaVA1.6 has not yet made their data publicly available}.

\item IT Dataset 2 (1.2M samples), used for evaluating the effect of \method-Adapter: To facilitate comparison with InstructBLIP and BLIVA models and showcase the \method-Adapter's effectiveness, we use a similar dataset with 1.2M samples. Where specific subsets are unavailable, we compensate by sampling additional examples from MSCOCO-based datasets with multiple questions and answers per image, ensuring a consistent 1.2M sample size.

\item IT Dataset 3 (8M samples), used for large-scale Instruction-tuning: We increase Dataset 2 to 8M image-instruction pairs, incorporating data from LVIS-Instruct4v \cite{lvis}, LAMM \cite{lamm}, Flickr30K \cite{flickr}, and ShareGPT4V \cite{sharegpt4v}. This larger and more diverse dataset allows assessing \method's generalization to broader visual concepts and instructions, and further highlight its efficiency in adapting to large-scale data. Our main results are reported using the model trained on this dataset. Details on how we mixed these datasets are given in the Appendix. 
\end{itemize}

\noindent{\textbf{Dataset balancing.}} We follow InstructBLIP balancing strategy to sample datasets with ratios proportional to the square root of their sizes (the numbers of training samples). Given $D$ datasets with sizes $\{N_1, N_2, \cdots, N_D \}$, we set the probability of a data sample being selected from a dataset $d$ during training is $p_d = \frac{\sqrt{N_d}}{\sum_{i=1}^D \sqrt{N_i}}$.  

\subsection{Implementation Details}

\subsubsection{Model architectures}
We adopt the ViT-G/14 architecture from EVA-CLIP~\cite{evaclip} as the vision encoder, that processes raw images of size 224x224. We extract features from its second to the last layer. As the LLM, we consider two distinct architectures: Vicuna7B/13B v1.5 (decoder-only), instruction-tuned from LLaMA2~\cite{llama}; and Flan-T5 XXL~\cite{flant5} (encoder-decoder), instruction-tuned from T5~\cite{t5}. Our Q-Former follows a design similar to BLIP2 and is initialized from the InstructBLIP model. It uses a set of 32 learnable query embeddings each with a dimension of 768.

\subsubsection{Training details}
We pretrain the projection layer for 5 epochs with a batch size of 32. During the instruction tuning stage, we employ a batch size of 16 with a maximum of 2M iterations, which roughly iterates over 4 epochs of the training data. For both training stages, we use the AdamW\cite{adamw} optimizer, with $\beta_1 = 0.9$, $\beta_2 = 0.999$, and a weight decay of 0.05. Additionally, we apply a linear warmup of the learning rate during the initial 1K steps, increasing from $10^{-8}$ to $10^{-5}$, followed by a cosine decay with a minimum learning rate of 0.

\subsubsection{MLLM zero-shot benchmarks}

We compare \method on a list of benchmarks for open-source MLLMs and the ones that are accessibly only via prediction-only APIs. We report results in MMMU~\cite{mmmu}, MME Perception ($\text{MME}^P$)\cite{mme}, MME Cognition ($\text{MME}^C$)\cite{mme}, MMBench (MMB)\cite{mmbench}, MM-Vet~\cite{mmvet}, HallusionBench (HallB)~\cite{hallb}, LLaVA-Bench In-the-Wild ($\text{LLaVA}^W$)~\cite{llava}, and SEED-Bench Image Part ($\text{SEED}^I$ )~\cite{seedbench}. MMMU benchmark is designed to evaluate multimodal models on multi-discipline tasks demanding college-level subject knowledge and deliberate reasoning.  $\text{MME}^P$ and $\text{MME}^C$ measure both perception and cognition abilities on a total of 14 subtasks. MMBench, contains approximately 3000 single choice questions covering 20 different ability dimensions, such as object localization and social reasoning. 
MM-Vet defines 6 core capabilities and examines the 16 integrations of interest derived from the capability combination. As the evaluation metrics, an LLM-based evaluator is used as `judge' for open-ended outputs. HallusionBench comprises 346 images paired with 1129 questions, all crafted by human experts to evaluate image-context reasoning with respect to visual hallucination. LLaVA-Bench In-the-Wild is a small dataset with a set of 24 images with 60 questions in total, including indoor/outdoor scenes, memes, paintings, sketches, etc. to evaluate the MLLM capability at challenging tasks and generalizability to novel domains. Lastly, SEED-Bench is a benchmark to text instruction following capabilities, consisting of 19K multiple choice questions with accurate human annotations which can objectively measure MLLMs performance without the need for human or LLM judge intervention. 

\subsubsection{Evaluation metrics} For consistency, we report official metrics computed using the standard implementations associated with each benchmark. 

\section{Results and Discussions}

\subsection{Zero-shot Performance on Vision-Language Tasks}
Table \ref{tab:zeroshot} shows comparisons on zero-shot performance on the 8 commonly-used MLLM benchmarks across various models, including the open-sourced ones and those who are accessible only via prediction APIs. Among the open-sourced ones, \method outperforms all baselines by a large margin on MMMU, $\text{MME}^P$, $\text{MME}^C$, MMBench, MM-Vet, and $\text{SEED}^I$.  
Considering the ones accessible only via prediction APIs, \method-Vicuna13B can outperform GPT4-V and Gemini Pro Vision on $\text{SEED}^I$ and Gemini Pro Vision on $\text{LLaVA}^W$ and $\text{MME}^P$.

\begin{table}[t]
\centering
\caption{Zero-shot performance of different MLLMs on  multimodal benchmarks. Input image resolutions, pretraining (PT) and instruction tuning (IT) dataset sizes are also shown. The best results among open-source modals are \textbf{bold} and the second best results are \underline{underlined}.} 
\label{tab:zeroshot} 
\begin{adjustbox}{width=0.98\textwidth}
\begin{tabular}{l|l|c|c|c|c|c|c|c|c|c|c|c}
\toprule
Method       & LLM         & Res. & PT   & IT    & MMMU    & $\text{MME}^P$     & $\text{MME}^C$    & MMB     & MM-Vet & HaLLB & LLaVA$^{W}$ & $\text{SEED}^I$  \\ \midrule
GPT4-V       & Unk         & Unk  & Unk  & Unk   & 56.8    &  \multicolumn{2}{c|}{1771.5}       & 77.0    & 67.7   & 65.8  & 93.1        & 69.1 \\
Gemini Pro Vision   & Unk         & Unk  & Unk  & Unk   & 47.9    &   1626.9      &   531.1     & 73.6    & 64.3   & 63.9  & 79.9        & 70.7 \\
OwenVL-Plus  & Unk         & Unk  & Unk  & Unk   & 46.5    & \multicolumn{2}{c|}{2229.8}   & 67.0    & 55.7   & 56.4  & 73.3 & 72.7 \\ \midrule \midrule
BLIP-2       & Vicuna-13B  & 224  & 129M & –    & 35.7    & 1293.8 & 290.0  &  –      & 22.4   &    –     & 38.1        & 46.4 \\
BLIVA        & Vicuna-7B    & 224  & 558K & 1.2M  &   27.3      &    813.4     &   224.6     &  60.1       &   30.4     &  33.9      &   80.1    &  59.4    \\
BLIVA        & Flan-T5 XXL  & 224  & 558K & 1.2M  &   –       &  1337.7   &   331.43     &    62.2     &    30.3    &   –     &     –        &   –    \\
InstructBLIP & Vicuna-7B   & 224  & 129M & 1.2M  & 30.6    &   –      &   –     & 33.9      & 26.2   & \underline{53.6}  & 60.9        & 53.4 \\
InstructBLIP & Vicuna-13B  & 224  & 129M & 1.2M  & 32.1  &  –       & –      &  –   & 25.6   &  –      & 58.2        &  –     \\
IDEFICS-80B  & LLaMA-65B   & 224  & 353M & 1M    & 24.0    &    –     & –      & 54.5    & 39.7   & 46.1  & –           & –    \\
Qwen-VL-Chat & Qwen-7B     & 448  & 1.4B & 50M  &   35.9      &   1487.6 & \underline{360.7}   &  61.8    &   47.3      &  \textbf{56.4 } &     –         & 65.4 \\
LLaVA-1.5    & Vicuna-7B   & 336  & 558K & 665K  &  –  & –  &  –  & 64.3    & 30.5   & –  & 63.4        & 58.6 \\
LLaVA-1.5    & Vicuna-13B  & 336  & 558K & 665K  & 36.4    & 1531 & 295 & 67.8    & 35.4   & 46.7  & 70.7        & 61.6 \\
LLaVA-1.6    & Vicuna-7B   & 336  & 558K & 760K  & 35.8 & 1519 & 332   &     67.4    & 43.9   &  –      & 81.6        & 70.2 \\
LLaVA-1.6    & Vicuna-13B  & 336  & 558K & 760K  & 36.2 & 1575 & 326   &    \underline{70.0}     & 48.4   &  –      &  \textbf{87.3 }      & 71.9 \\ \midrule
\textbf{\method} (ours)   & Vicuna-7B   & 224  & 558K & 8M &  \underline{38.8}     &  \underline{1590.6} & 358.4 &   67.1      & 47.3 &   48.3    & 79.2        & 68.2 \\
\textbf{\method } (ours)  & Vicuna-13B  & 224  & 558K & 8M  &     \textbf{41.2}   & \textbf{1643.2}    &  \textbf{369.5}     &   \textbf{71.2 }   &    \textbf{55.1}    &   49.2   &   \underline{86.8}          &   \underline{72.5}   \\
\textbf{\method} (ours)  & Flan-T5 XXL & 224  & 558K & 8M  &   38.1      &  1521.3       &  361.4      &    65.5     &      45.2  &  51.3     &   76.4          &   \textbf{73.5 }  \\ \bottomrule 
\end{tabular}
\end{adjustbox}
\end{table}

Table \ref{tab:dataset_size} compares \method trained on different dataset sizes, each benchmarked against a corresponding baseline with similar LLM architecture and size. Comparing the models with Flan-T5 XXL backbone, \method trained with 1.2M IT samples, shows superior performance on $\text{MME}^P$, MMBench, MM-Vet, LLaVA$^{W}$ and $\text{SEED}^I$ benchmarks. When trained with 8M IT samples, \method consistently outperforms BLIVA and InstructBLIP on all the benchmarks. This highlights that \method can effectively take advantage of more instruction tuning samples. 

Using a similar instruction-tuning dataset as BLIVA and InstructBLIP, \method achieves higher performance on 6 out of 8 benchmarks. On HallB, InstructBLIP's larger pretraining dataset of 129M samples, contributes to its advantage, but \method remains competitive. On LLaVA$^{W}$, LLaVA and BLIVA outperform \method with margins of 2.4 and 0.9 points respectively, while other models lag by at least 15 points. On $\text{SEED}^I$, \method surpasses BLIVA and InstructBLIP but falls short of LLaVA, likely due to LLaVA's higher-resolution image encoder which benefits this visually-rich benchmark.
 
Intriguingly, when comparing \method-Vicuna-7B across dataset sizes (1.2M and 8M) with LLaVA-7B-1.6, we see that larger training data (8M) enables \method to outperform a model with more parameters. Further, increasing \method's LLM backbone (Vicuna 13B) widens the performance gap between \method and LLaVA, emphasizing the effectiveness of our modality alignment module in leveraging the LLM's existing capabilities.

\begin{table}[htbp]
\centering
\caption{Comparison between \method models trained on different pretraining (PT) and instruction tuning (IT) dataset sizes and corresponding baselines with similar LLM backbones. The total number of trainable parameters and the input image resolutions are also shown. The best results in each LLM family of models are \textbf{bold}.} 
\label{tab:dataset_size} 
\begin{adjustbox}{width=0.98\textwidth}
\begin{tabular}{l|l|c|c|c|c|c|c|c|c|c|c|c|c}
\toprule
Method       & LLM  &  \begin{tabular}[c]{@{}l@{}}\#params\end{tabular}       & Res. & PT   & IT    & MMMU    & $\text{MME}^P$     & $\text{MME}^C$    & MMB     & MM-Vet & HaLLB & LLaVA$^{W}$ & $\text{SEED}^I$  \\ \midrule
InstructBLIP & Flan-T5 XXL & 188M &  224  & 129M & 1.2M  &   35.7      &    1212.8     &  291.8      &    –         &   25.6     &     –      &      58.2       &  52.7    \\
BLIVA        & Flan-T5 XXL  & 194.61M & 224  & 558K & 1.2M  &   –       &  1337.7   &   331.4    &    62.2     &    30.3    &   –      &     –        &   –    \\
\textbf{\method}(ours)    & Flan-T5 XXL & 199.85M & 224  & 558K & 1.2M  &    33.1     &     1380.6    &     329.2   &  63.1       &   34.7     &    40.1   &        70.2     &  62.7    \\ 
\textbf{\method} (ours)  & Flan-T5 XXL & 199.85M  & 224  & 558K & 8M  &  \textbf{ 38.1}      &  \textbf{1521.3 }      &  \textbf{361.4}      &   \textbf{ 65.5  }   &     \textbf{ 45.2}  &  \textbf{51.3}     &  \textbf{ 76.4  }        &   \textbf{73.5 } \\ \midrule
InstructBLIP & Vicuna-7B   & 188M & 224  & 129M & 1.2M  & 30.6    &   –      &   –     & 33.9      & 26.2   & \textbf{53.6}  & 60.9        & 53.4 \\
BLIVA        & Vicuna-7B   & 194.61M & 224  & 558K & 1.2M  &   27.3      &    813.4     &   224.6     &  60.1       &   30.4     &  33.9      &   80.1    &  59.4    \\
LLaVA-1.6    & Vicuna-7B & 7B  & 336  & 558K & 760K  & 35.8 & 1519 & 332   &     \textbf{67.4}    & 43.9   &  –      & \textbf{81.6}        & \textbf{70.2} \\
\textbf{\method}(ours)      & Vicuna-7B &  199.85M & 224. & 558K & 1.2M  &  32.4  &   1170.4	& 246.1    &     62.8    &  32.6      &  42.3      &     65.2       &   60.3   \\
\textbf{\method} (ours)   & Vicuna-7B   &  199.85M  & 224  & 558K & 8M &   \textbf{38.8}     &  \textbf{1590.6} & \textbf{358.4} &   67.1      & \textbf{47.3}   &   48.3    & 79.2        & 68.2 \\ \midrule
LLaVA-1.6    & Vicuna-13B  & 13B  & 336  & 558K & 760K  & 36.2 & 1575 & 326   &    70.0     & 48.4   &  –      & \textbf{87.3}        & 71.9 \\ 
\textbf{\method } (ours)  & Vicuna-13B  &  203.39M & 224  & 558K & 8M  &     \textbf{41.2}   & \textbf{1643.2}    &  \textbf{369.5}      &    \textbf{71.2}     &    \textbf{55.1}    &   49.2   &   86.8          &   \textbf{72.5}   \\  \bottomrule
\end{tabular}
\end{adjustbox}
\end{table}

\begin{center}
  \begin{minipage}[t]{0.48\textwidth}
    \centering
    \captionof{table}{ScienceQA-Image results: zero-shot vs. fine-tuned performance.}    
    \begin{adjustbox}{width=\textwidth}
    \begin{tabular}{@{}llc@{}}
\toprule
Methods                                     & \begin{tabular}[c]{@{}l@{}}Trainable\\ params\end{tabular} & Accuracy (\%) \\ \midrule
\multicolumn{3}{c}{\textbf{Zero-Shot Performance}}                                                                           \\ \midrule
LLaVA 1.6-7B    &  7B   &  	70.1  \\
InstructBLIP-Vicuna-7B &   188M &   60.5            \\ 
BLIVA-Vicuna-7B        &  194.61M     &   57.3       \\\midrule
\method-Vicuna-7B (ours) &  199.85M    &     61.2      \\ \midrule
\multicolumn{3}{c}{\textbf{Supervised Fine-tuning}}                                                                          \\ \midrule
        Mutimodal-T-SciQLarge & 738M   & 96.2 \\
        MC-CoT-F-Large       & 738M    & 94.9 \\
        LLaMA-Adapter         & 1.8M   & 85.2 \\
        LaVIN-7B              & 3.8M   & 89.4 \\
        LaVIN-13B             & 5.4M   & 90.8 \\
        PILL-7B               & 45 M   & 91.2 \\ \midrule
        \textbf{\method-Vicuna-7B} (ours)    & 5.24M &  93.2\\ \bottomrule
    \end{tabular}
    \end{adjustbox}
    \label{tab:sqa}
  \end{minipage}%
  \hfill
  \begin{minipage}[t]{0.48\textwidth}
    \centering
    \captionof{table}{AI2D results: zero-shot vs. fine-tuned performance. Note that Qwen-VL-Chat includes AI2D in pretraining, while BLIVA, InstructBLIP and \method do not.}    
    \begin{adjustbox}{width=\textwidth}
    \begin{tabular}{@{}llc@{}}
    \toprule
    Methods                                     & \begin{tabular}[c]{@{}l@{}}Trainable\\ params\end{tabular} & Accuracy (\%) \\ \midrule
    \multicolumn{3}{c}{\textbf{Zero-Shot Performance}}                                                                           \\ \midrule
    \multicolumn{1}{l|}{Qwen-VL-Chat$^*$}              & \multicolumn{1}{l}{}        9.6B        &   57.7       \\
    \multicolumn{1}{l|}{InstructBLIP-Vicuna-7B} & \multicolumn{1}{l}{}        188M      &   36.1       \\ 
    \multicolumn{1}{l|}{BLIVA-Vicuna-7B}               & \multicolumn{1}{l}{}        194.61M      &    38.2      \\ \midrule
    \multicolumn{1}{l|}{CROME-Vicuna-7B (ours)} & \multicolumn{1}{l}{}        199.85M   &  39.1        \\ \midrule
    \multicolumn{3}{c}{\textbf{Supervised Fine-tuning}}                                                                          \\ \midrule
    \multicolumn{1}{l|}{InstructBLIP-Vicuna-7B} & \multicolumn{1}{l}{}        188M   &  65.0        \\
    \multicolumn{1}{l|}{BLIVA-Vicuna-7B}        & \multicolumn{1}{l}{}        194.61M   &  69.2         \\ \midrule
    \multicolumn{1}{l|}{\textbf{\method-Vicuna-7B} (ours)} & \multicolumn{1}{l}{}        5.24M  & 75.3         \\ \bottomrule
    \end{tabular}
    \end{adjustbox}
    \label{tab:ai2d}
  \end{minipage}
\end{center}

\subsection{Task-Specific Fine-tuning}
We evaluate \method's task-specific fine-tuning capabilities considering small-scale labeled datasets for two tasks: 
(i) ScienceQA-Image ($\text{SQA}^I$)~\cite{sqa}, containing elementary and high school science curricula and 
(ii) AI2D~\cite{ai2d}, which covers diagrams from grade school science. 
These datasets are specifically chosen to assess \method's adaptation to unseen tasks, as neither they nor similar scientific or math VQA content are included in our training data. 

We initialize \method from the instruction-tuned model (for which we report the zero-shot performance), and selectively train only the \method-Adapter parameters during fine-tuning. Training details are provided in the Appendix.
As they are both multiple-choice outputting tasks, we use accuracy as the evaluation metric.

As shown in \ref{tab:sqa}, \method achieves an impressive 93.2\% accuracy on $\text{SQA}^I$, significantly improving its zero-shot performance by 32\%. Despite lacking the data augmentation via prompting in chain-of-thought (CoT) baselines which have multiple stages of training to create rationals as additional context ~\cite{tsci,mmcot} and training with more parameters, \method demonstrates competitive performance. Moreover, \method outperforms other adapter-based approaches~\cite{lavin,pill} in accuracy while utilizing much lower amount of trainable parameters. Notably, the proposed adapter's pre-LLM placement distinguishes it from LaVIN and PILL, which incorporate adapters within the Transformer layers of LLaMA-based models. 

Table \ref{tab:ai2d} presents zero-shot and fine-tuned performance on the AI2D dataset for \method, InstructBLIP, and BLIVA. We include Qwen-VL-Chat's reported results as a baseline, noting their use of AI2D in pretraining, unlike the other models.  At zero-shot, \method surpasses BLIVA by 0.9\% and InstructBLIP by 3\%. However, without prior exposure to science-related VQA, none of the models achieve high accuracy. Despite extensive training (9.6B parameter updates) and including AI2D in its pretraining, Qwen-VL-Chat's performance remains at 57.7\% while fine-tuning \method with its \method-Adapter significantly boosts accuracy on this dataset, achieving a 36.2\% improvement over its zero-shot performance. Compared to InstructBLIP and BLIVA, which retrain their Q-Former and projection layers during fine-tuning, \method achieves 10.3\% and 6.1\% higher accuracy, respectively, while training only 2.5\% of their parameters. This demonstrates the remarkable efficiency of our adapter-based fine-tuning approach.

\subsection{Ablation Studies}
To gain a deeper understanding of \method, we conduct ablation studies focusing on its key components.  We use the closely related BLIVA model as our baseline for comparison. For evaluation, we employ a zero-shot MLLM benchmark ($\text{MM-Vet}$) and the $\text{SQA}^I$ dataset which highlights our efficient fine-tuning strategy.  Table \ref{tab:ablation} summarizes two types of ablations performed at model, \method-Adapter in particular, and data levels.

\subsubsection{Architecture:}
We first ablate the \method-Adapter to highlight its effect in our model. This ablation essentially reduces the architecture to BLIVA~\cite{bliva} for which the results are shown in the first row of \ref{tab:ablation}. On MM-Vet, BLIVA achieves the score of 30.4 while adding the \method-Adapter with its gating mechanism obtains 32.6 score (ablation \#3) using similar pretraining and instruction-tuning data, LLM, and vision encoder. On $\text{SQA}^I$, \method achieves 61.2\% zero-shot accuracy while BLIVA reaches 57.3\% under similar conditions. 

Ablation \#2 shows the effect of the gating mechanism in \method-Adapter. For this ablation, we remove the gating layer of the adapter in the down projection unit which results in simplifying \ref{eq:gating} to $\textbf{z}(x) = \text{SiLU}(x \textbf{W}_d)$. We repeat pretraining and instruction-tuning \method with Dataset \#2. Notably, the performance on MM-Vet and $\text{SQA}^I$ downgrade by 1.7 and 3, respectively, which shows the effect of the component-wise product of the two linear layers in the down-project unit.

\begin{table}[t]
\centering
\caption{Ablation studies for \method framework using \method-Vicuna-7B on MM-Vet and ScienceQA-Image datasets}
\label{tab:ablation}
\begin{adjustbox}{width=0.95\textwidth}
\begin{tabular}{@{}c|c|c|c|c|c|c||l|l@{}}
\toprule
\# & 
BLIVA~\cite{bliva} &   
\begin{tabular}[c]{@{}l@{}}\method-\\Adapter\end{tabular} &
\begin{tabular}[c]{@{}l@{}}Gating\\Mechanism\end{tabular} &
\begin{tabular}[c]{@{}l@{}}Pretraining\end{tabular} &
\begin{tabular}[c]{@{}l@{}}Additional\\ IT Data\end{tabular} &
\begin{tabular}[c]{@{}l@{}}Task-specific \\ IT Strategy\end{tabular} &
  MM-Vet & 
  $\text{SQA}^I$ \\ \midrule

1 & \checkmark &           &           &  \checkmark  &       &    &  30.4 (+0) & 57.3\% (+0)\\
2 & \checkmark          & \checkmark &     & \checkmark      &           &  & 28.7 (-1.7)   &  54.3\% (-3.0) \\
3 & \checkmark          & \checkmark & \checkmark &  \checkmark  &       &  & 32.6 (+2.2)  & 58.2\% (+0.9) \\ \midrule
4 & \checkmark          & \checkmark & \checkmark &  &  & & 31.8 (+1.4)   &  57.6\% (+0.3) \\ 
5 & \checkmark          & \checkmark & \checkmark & \checkmark &  \checkmark &  & 47.3 (+16.9)  &  61.2\% (+3.9) \\ 
6 & \checkmark          & & & \checkmark &  &\checkmark  &  ~~~~~~~ –  & 84.4\% (+27.1) \\
7 & \checkmark         & \checkmark & \checkmark & \checkmark & \checkmark & \checkmark &  ~~~~~~~ –  & 93.2\%  (+35.9) \\\bottomrule
\end{tabular}
\end{adjustbox}
\end{table}

\subsubsection{Data and training:}
Table \ref{tab:ablation}, row \#4, shows the effect of pretraining data in \method which is not significant compared to other components (1.3 score difference on MM-Vet and 0.3\% improvement on $\text{SQA}^I$). 

Table \ref{tab:ablation}, row \#5, demonstrates the significant impact of expanding the instruction-tuning dataset from 1.2M to 8M samples. This leads to a 16.9 point improvement in \method's zero-shot performance on MM-Vet and a 3.9\% increase on $\text{SQA}^I$. 

Ablation \#6 isolates the effects of the \method-Adapter and additional instruction-tuning data, yielding 84.4\% accuracy, 8.8\% lower compared to the full \method approach (row \#7). This highlights the importance of the cross-modal gated adapter, supervised fine-tuning with it, and the better pretrained checkpoint trained on larger-scale instruction-tuning data.

\subsection{Qualitative Results} \label{sec:qualitative_examples}
In addition to~\ref{fig:teaser} 
we include additional case examples from a variety of tasks performed by \method and other models in the Appendix.

\section{Limitations}
\method is designed to follow instructions and answer questions about images, outputting textual responses. Similar to other autoregressive LLMs, there is a potential for occasional inaccuracies or inconsistencies in its reasoning. While we prioritize the use of publicly available, curated data during training, it is important to acknowledge the potential for biases within these datasets that might influence \method's responses. Aside from these considerations, we are optimistic that \method will contribute to streamlined training and adaptations of MLLMs. Addressing potential challenges related to hallucination mitigation and improved grounding in MLLMs remain important future work directions.
 
\section{Conclusion}

In this paper, we introduce \method, a novel vision-language instruction tuning framework designed for parameter-efficient multimodal learning and task-specific adaptation. 
\method features a lightweight, gated cross-modal adapter that fuses visual and textual representations before they are input into the LLM, which is proposed to kept frozen. 
This design promotes efficient cross-modal understanding while minimizing computational costs by avoiding extensive LLM retraining. 
Additionally, we highlight the potential for parameter-efficient fine-tuning of \method: by training only the adapter's small number of parameters (O(M)-O(10M)), we demonstrate outperforming the state-of-the-art approaches on two downstream tasks. 
Furthermore, \method achieves superior zero-shot performance on commonly-used MLLM benchmarks. 
We leave extending \method to incorporate other modalities (\textit{e.g.} audio and video), investigating broader task-specific adaptation, and further optimizing  \method's cross-modal adapter architecture to future work.

\newpage

%
%
\bibliographystyle{unsrt}
\bibliography{neurips_2024}

\clearpage
\appendix

{
\begin{center}
\maketitle
\Large 
\textbf{Appendix} 
\\
\end{center}
\vspace{2pt}
} 

\section{Implementation Details}

\subsection{Image pre-processing}
We pre-process the images using random crops, resizing to 224 $\times$ 224 with an interpolation method of Bicubic, horizontal flips, converting them to tensor format, and normalizing them using mean = (0.48145466, 0.4578275, 0.40821073) and standard deviation = (0.26862954, 0.26130258, 0.27577711). During evaluation, we only used image resizing, converting to tensor format, and normalizing using the same mean and standard deviation values.

\subsection{Training details of task-specific fine-tuning}
On both ScienceQA-Image and AI2D datasets, we employ a batch size of 16 and used the AdamW\cite{adamw} optimizer, with $\beta_1 = 0.9$, $\beta_2 = 0.999$, and a weight decay of 0.05. Additionally, we apply a linear warmup of the learning rate during the initial 1K steps, increasing from $10^{-8}$ to $10^{-4}$, followed by a cosine decay with a minimum learning rate of 0. 

\section{Training Datasets}
Table \ref{tab:datasets} lists the datasets used in pretraining and instruction tuning of \method. With the recent release of publicly available instruction tuning datasets such as ShareGPT4V, LAMM, and LVIS-Instruct4V, we were able to increase the dataset size to 8M image-text pairs. It should be noted that when multiple questions/instructions were available for the same image, we used up to 4 of them. Moreover, duplicated data across these datasets which had the same image and text were filtered during our cleaning process. 

\begin{table}[ht]
\centering
\caption{Datasets used in pretraing and instruction tuning \method}
\label{tab:datasets}
\begin{tabular}{|l|l|}
\hline
Phase        & Dataset                                                       \\ \hline
Pretraining & LLaVA 558K (filtered image-text pairs from LAION, CC-3M, SBU) \\ \hline
IT w/ D1 (665K) & \begin{tabular}[c]{@{}l@{}}LLaVA 158K, ShareGPT,  VQAv2, OKVQA, A-OKVQA, \\ OCR-VQA, TextCap,  GQA, RefCOCO,  VG\end{tabular} \\ \hline
IT w/ D2 (1.2M) & \begin{tabular}[c]{@{}l@{}}LLaVA-Instruct 150K, VQAv2, OKVQA, A-OKVQA, \\ OCR-VQA, TextCap, MSCOCO\end{tabular}               \\ \hline
IT w/ D3 (8M)   & \begin{tabular}[c]{@{}l@{}}LLaVA-Instruct 150K, VQAv2, OKVQA, A-OKVQA, \\ OCR-VQA, MSCOCO, LVIS-Instruct4V, LAMM, \\ ShareGPT4V, Flickr30K\end{tabular} \\ \hline
\end{tabular}
\end{table}

\subsection{Analysis on \method-Adapter bottleneck dimension}
Table \ref{tab:m} shows how we chose the $m$ value for the hidden dimension of our \method-Adapter. We only did this experiment on the \method-Vicuna7B model and used the chose $m$ value for other variants of \method-Vicuna13B  and \method-Flan T5 XXL. We used two datasets, one from the zero-shot MLLM benchmarks (MM-Vet) as well as one of the datasets we used for supervised finetuning adaptation (ScienceQA-Image). It is common to use a significantly smaller dimension for the bottleneck with respect to the input size ($m << d (4096) $). We selected 4 values (64, 128, 256, 512) and pre-trained and instruction-tuned four models with them. Results on both datasets confirmed the choice of $256$ for our model.

\begin{table}[]
\centering
\caption{Analysis on choosing $m$, the bottleneck dimension in \method-Adapter using one dataset from zero-shot MLLM benchmarks as well as ScienceQA Image dataset.}
\label{tab:m}
\begin{tabular}{l|ll}
\hline
$m$           & MM-Vet & SQA-IMG \\ \hline
64          &    46.1    &    89.1     \\
128         &    46.8    &    91.9     \\
256         &    47.3    &    93.2     \\
512         &    46.    &    92.3     \\ \hline
\end{tabular}
\end{table}

\section{Qualitative Results}
\begin{figure}[t]
\begin{center}
  \includegraphics[scale=0.45]{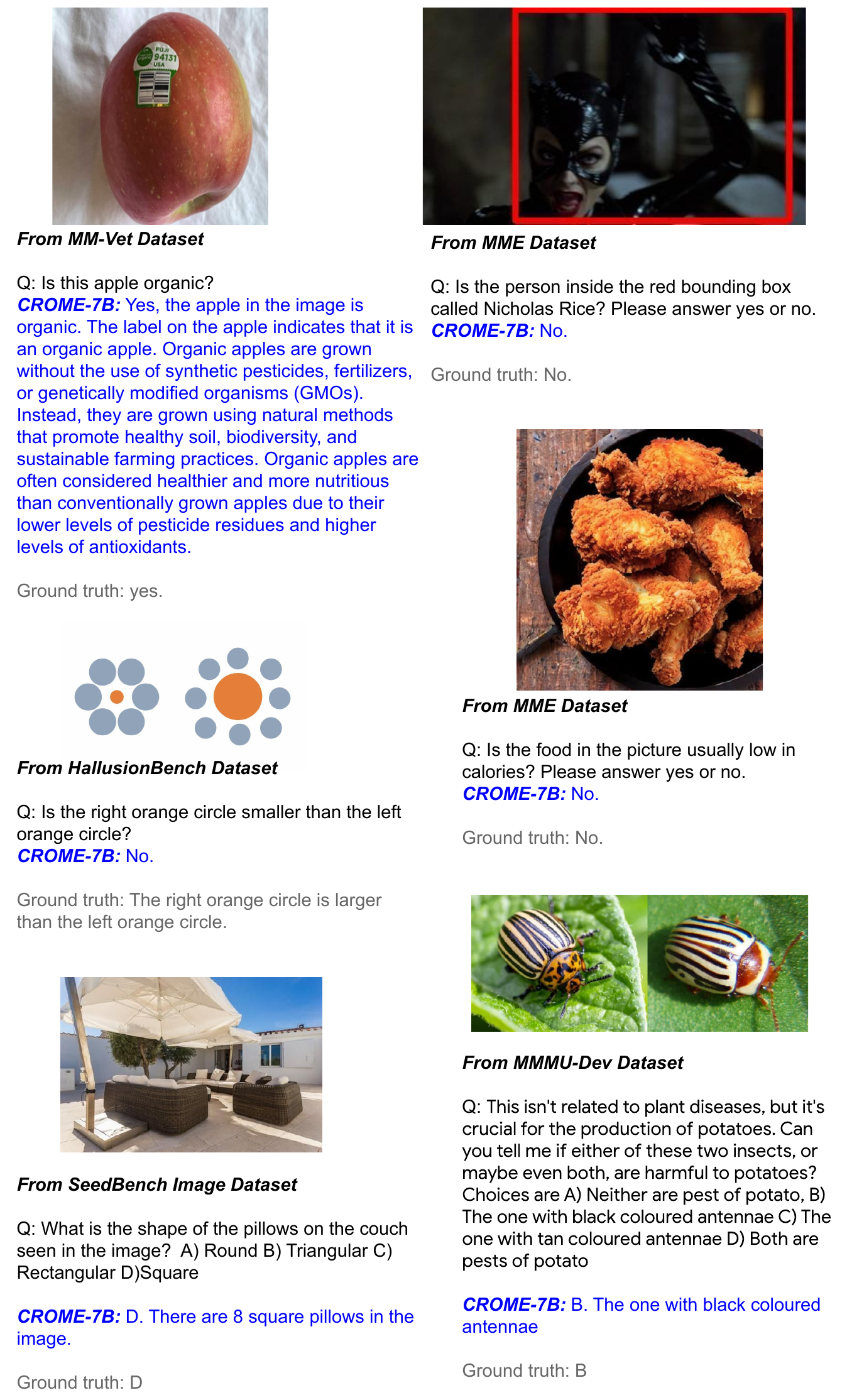}
\end{center}
  \caption{Qualitative examples from various zero-shot MLLM benchmarks we have evaluated \method on.}
\label{fig:qualitative}
\end{figure}

Figure \ref{fig:qualitative} shows some qualitative results from \method on various test samples. 

\end{document}